\documentclass[review]{elsarticle}
\usepackage{graphicx}
\usepackage{url}
\usepackage{booktabs}
\usepackage{lscape}
\usepackage{color}
\usepackage{rotating}
\usepackage{times}
\usepackage{latexsym}
\usepackage{multirow}
\usepackage{enumitem}
\usepackage{listings}
\usepackage{float}

\usepackage{amsthm}
\usepackage{amsmath}
\usepackage{longtable}
\usepackage{booktabs}
\usepackage{lscape}
\usepackage{rotating}
\usepackage{amsmath,amssymb,amsfonts}
\usepackage{times}
\usepackage{latexsym}
\usepackage{algorithmic}
\usepackage{textcomp}
\usepackage{url}
\usepackage[linesnumbered,ruled,vlined]{algorithm2e}
\usepackage{graphicx}
\usepackage{booktabs}
\usepackage{array}
\usepackage{url}
\graphicspath{{./}}
\begin{document}

\begin{frontmatter}

\title{An unsupervised and customizable misspelling generator for mining noisy health-related text sources}

\author{Abeed Sarker\fnref{myfootnotex}}
\address{Department of Biostatistics, Epidemiology and Informatics, Perelman School of Medicine, University of Pennsylvania}
\fntext[myfootnotex]{Corresponding Author. Address: 423 Guardian Drive, Philadelphia, Pennsylvania 19104. Phone: 1-602-474-6203. Email: abeed@pennmedicine.upenn.edu}


\author{Graciela Gonzalez-Hernandez}
\address{Department of Biostatistics, Epidemiology and Informatics, Perelman School of Medicine, University of Pennsylvania}

\begin{abstract}\\
 \textbf{Background}\\
Data collection and extraction from noisy text sources such as social media typically rely on keyword-based searching/listening. However, health-related terms are often misspelled in such noisy text sources due to their complex morphology, resulting in exclusion of relevant data for studies. In this paper, we present a customizable data-centric system that automatically generates common misspellings for complex health-related terms.

\textbf{Materials and Methods}\\
The spelling variant generator relies on a dense vector model learned from large unlabeled text, which is used to find semantically close terms to the original/seed keyword, followed by the filtering of terms that are lexically dissimilar beyond a given threshold. The process is executed recursively, converging when no new terms similar (lexically and semantically) to the seed keyword are found. Weighting of intra-word character sequence similarities allows further problem-specific customization of the system.

\textbf{Results}\\
On a dataset prepared for this study, our system outperforms the current state-of-the-art for medication name variant generation with best $F_1-score$ of 0.69 and $F_\frac{1}{4}-score$ of 0.78. Extrinsic evaluation of the system on a set of cancer-related terms showed an increase of over 67\% in retrieval rate from Twitter posts when the generated variants are included.

\textbf{Discussion}\\
Our proposed spelling variant generator has several advantages over the current state-of-the-art and other types of variant generators---(i) it is capable of filtering out lexically similar but semantically dissimilar terms, (ii) the number of variants generated is low as many low-frequency and ambiguous misspellings are filtered out, and (iii) the system is fully automatic, customizable and easily executable. While the base system is fully unsupervised, we show how supervision maybe employed to adjust weights for task-specific customization.  

\textbf{Conclusion}\\
The performance and significant relative simplicity of our proposed approach makes it a much needed misspelling generation resource for health-related text mining from noisy sources. The source code for the system has been made publicly available for research purposes.\\

\end{abstract}

\end{frontmatter}
\newpage
\section{Introduction}

Massive amounts of medical knowledge are encapsulated in the form of noisy, unstructured text in sources such as social media and clinical notes. As the volume of electronically available unstructured text continues to surge, the knowledge contained within these resources is becoming increasingly useful for solving many research problems. Over the recent years, research on utilizing data garnered from noisy text sources such as social media has seen steady growth in the broader health domain. Research in the sub-domain of public health, for example, has focused on devising methods that can make population-level estimates from social media data \cite{brownstein09, paul11,sinnenberg17} for tasks such as virus/flu outbreak surveillance \cite{broniatowski13,kagashe17}, pharmacovigilance \cite{sarker15,cocos17} and drug use/abuse \cite{sarker16tox,kazemi17}, to name a few. Similarly, unstructured data (\emph{i.e.}, free text) from clinical notes has been used for notable tasks such as predicting suicide \cite{poulin14}, identifying risk factors for targeted populations \cite{khalifa15} and identifying relations between known clinical entities \cite{luo17}. The adoption of noisy text sources in health research has been driven largely by the advances in technology and data science methods. However, the use of such data is not without its challenges from the perspective of natural language processing (NLP). One key challenge is the inevitable presence of misspellings in human authored texts \cite{keselman12}. The problem is exacerbated for medical text as domain-specific terms are typically difficult to spell \cite{zhou15}. In the recent past, studies have attempted to apply preprocessing techniques to attempt to correct misspellings prior to the application of downstream NLP and machine learning methods such as information extraction and classification \cite{lai15}. Such preprocessing methods for spelling correction are, however, only useful once the required data have already been collected, and they do not aid in the initial step of data collection or concept detection during information retrieval/searching. In this paper, we describe a data-centric system that addresses this issue by generating potential common misspellings given the original terms. We describe the development and evaluation of the system on text from social media, specifically Twitter, which is one of the noisiest sources of free text.

The first step in incorporating social media data for operational and research tasks involves designing data extraction/collection strategies, which are typically reliant on keyword-based searches (\emph{e.g.}, \cite{alvaro15}). Twitter is a commonly used resource for social media based NLP research within the health domain, and it provides a public Application Programming Interface (API) that can be queried using predefined keywords. However, as mentioned, complex medical terms, such as medication and disease names, are often misspelled by users on Twitter \cite{karimi15}. For example, when preparing data for this study, we found 46 common misspellings for the medication name \emph{clonazepam} including \emph{klonazapam}, \emph{colnazepam} and \emph{clonazpam}. Presence of such misspellings is problematic from the perspective of data collection and concept detection, as all the possible misspellings are not known \emph{a priori}. Using only the correctly spelled keywords leads to loss of potentially important data, particularly for terms orc concepts with difficult spellings and morphology. Some misspellings also occur more frequently than others, meaning that they are more important for data collection and concept detection.

Despite the importance of the task of automatic misspelling generation, it has received little research attention from the NLP community. This task can be viewed as the opposite of spelling correction. In spelling correction, the goal is to detect out-of-vocabulary terms and map them to a finite set of in-vocabulary terms \cite{baldwin15}. In contrast to misspelling generation, automatic spelling correction methods have been studied for decades, with early approaches employing the noisy channel framework \cite{church91,brill00}, which require gold-standard lexicons that can be used to learn transformations (generally character edits) required for correcting misspellings. Two categories of approaches have been employed in the recent past for performing spelling correction---lexicon-based and language model-based \cite{han13}. The former category of approaches are rule-based and require the building of extensive lookup tables which map out-of-vocabulary terms to in-vocabulary terms. Such approaches are useful if the target vocabularies are small and the misspellings are unambiguous. Building extensive lookup tables, however, is infeasible for most tasks as the total number of possible misspellings can be very large and the distributions of the misspellings are not known beforehand. For the same reason, misspelling generation from manually built lexicons is not feasible in most cases. Unlike lexicon-based approaches, language model-based approaches are capable of disambiguating potential in-vocabulary candidates by deriving likelihood measures based on the context of the out-of-vocabulary terms in question. Unfortunately, language model-based spelling correction approaches, on their own, have not been very successful \cite{sarker17lexnorm}, and recent, high-performance approaches have been hybrid in nature---combining rules with language models learned from noisy data \cite{berend15}. While it is not possible to reverse and apply the successful spelling correction methods for the task of misspelling generation, our approach is inspired by the same principles---we combine rules with a distributed language model learned from a large, unlabeled, domain-specific dataset to accomplish the task.


Some studies have attempted to resolve the misspelling generation problem by manually curating spelling variants or synthetically generating them \cite{sloane15}. A na\"ive approach to addressing the problem is to generate all variants that are lexically similar to the original term. Lexical similarity/dissimilarity between two terms can be measured in terms of edit distance---the minimum number of operations required to transform one term to the other. Thus, the na\"ive approach would generate all character combinations that are within a given edit distance threshold (\emph{e.g.}, 2). This results in the generation of too many character combinations, even at very low thresholds, particularly if the original keyword is long. Incorporating variants that do not represent misspellings for data collection may result in the retrieval of excess noise or face other operational obstacles such as API limits. Twitter, for example, at the time of publication, has a limit of 400 keywords per API key. Due to these reasons, it is important to constrain the number of variants generated to highly precise ones. 

The current state-of-the-art system by Pimpalkhute \emph{et al.} \cite{pimpalkhute14} attempts to address these problems by incorporating the notion of phonetic similarity. In their approach, the authors limit the number of variants to those that are phonetically close to the original keyword. The authors also use the Google Search API to identify the most frequent misspellings occurring in web search queries. While the approach does significantly limit the number of variants, it faces several drawbacks. First, the variants are limited to only edit distance 1, resulting in the loss of many genuine misspellings that are farther from the original keyword in terms of edit distance. Second, there may be some terms that fit the edit and phonetic distance constraints, but are semantically very dissimilar. For example, for the medication name \emph{activase}, one potential misspelling that fits all the abovementioned criteria is \emph{activate}. Incorporating the term \emph{activate} in data collection or searching results in very noisy data, as this variant occurs at a much higher frequency in free text compared to the medication name and its genuine misspellings. Third, the proposed approach is semi-automatic, and several steps, including manual ones, are required to generate the spelling variants for a single keyword. This makes the process of variant generation very cumbersome when hundreds of keywords need to be processed. Fourth, the approach is dependent on external sources and APIs, such as the Google Search API, which have their own limits. On top of these drawbacks, the number of irrelevant misspellings generated per keyword by this approach is still quite high. 

We address the abovementioned drawbacks of current approaches by proposing a simple recursive method for generating small sets of highly precise misspellings for restricted-domain keywords. The proposed method relies on a distributed word vector model \cite{mikolov13} learned from large domain-specific text to first identify a set of semantically similar candidate misspellings for a given keyword. Semantic similarity is computed using the \emph{cosine similarity} measure over the vector representations of the terms. Next, a lexical similarity filter using \emph{Levenshtein ratio} is applied to exclude lexically dissimilar terms beyond a given threshold. These two steps are applied recursively to each detected misspelling for a keyword until no further misspellings are found. While the default system is unsupervised, we show that supervision can be incorporated into the system through a weight optimization method for intra-word character sequence similarities, allowing for problem-specific tuning of the system. We evaluated our method intrinsically on the problem of medication name misspelling generation and compared the performances of two settings of the system to the phonetic filter based state-of-the-art. On a manually prepared evaluation set, our approach obtained best $F_1$-$score$ of $0.69$ and $F_{\frac{1}{4}}$-$score$ of $0.78$, significantly outperforming the competing system. We performed extrinsic evaluation by comparing the data collection rates with and without spelling variants for a set of cancer terms, and observed an increase of over 67\% when the misspellings are included. We have made the system and the source code publicly available for the research community.\footnote{The source code is available at \url{https://bitbucket.org/asarker/qmisspell}.}

\section{Materials and methods}

Our automatic misspelling generation approach can be viewed as a two-step process. In the first step, word or phrase level dense vectors (embeddings) are generated from a large unlabeled dataset. These embeddings capture semantic relatedness between terms occurring in the unlabeled text. In the second step, the embedding model is used as input in a recursive algorithm that selectively filters out unlikely misspelling candidates. We now discuss these two steps in detail.

\subsection{Dense vector language model}
Dense word and phrase vector models have become very popular for NLP research in recent years due to their ability to very accurately encode semantic information. To learn such language models, large unlabeled data sets are required, of which there is an abundance in social media and over the Internet in general. The word2vec algorithm \cite{mikolov13} is among the most common approaches in generating such dense vector models. This algorithm learns vector representations for terms that occur more frequently than a given threshold (\textit{e.g.}, 10) by training a shallow neural network. Terms that occur in similar contexts result in similar vectors, and, thus, are close to each other in semantic space. Intuitively, since common misspellings of terms and their original forms appear in similar contexts, their vectors are likely to be similar as well.

\begin{figure}[htbp] 
	\centering
	\scalebox{0.35}
	{\includegraphics{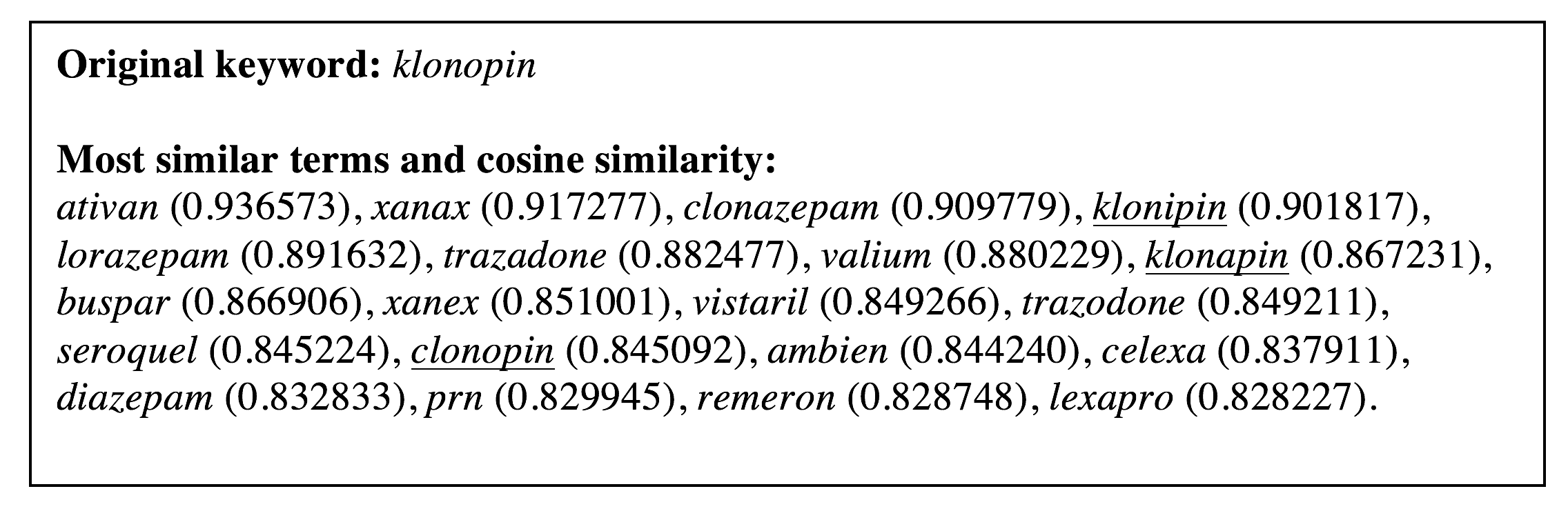}}
	\caption{The 20 most similar terms for \textit{klonopin} in our dense vector model and their cosine similarities. Misspellings of the original term are underlined.}
	\label{densevectorexample}
\end{figure}

Our misspelling generation process requires a word vector model generated from the relevant domain-specific text, such that the model contains sufficient numbers of misspellings and their in-vocabulary forms. We used our own publicly released model from past work \cite{sarker2017dib}, which was generated from medication-related chatter from social media.\footnote{The model is available at: \url{https://data.mendeley.com/datasets/dwr4xn8kcv/3}. We used the 2.2 GB n-gram model for the experiments reported in this paper.} Each word or phrase in the model is represented via a 400 dimensional vector. The model was learned using a context window size of 5 and only n-grams occurring more than 40 times were included. Figure \ref{densevectorexample} shows the 20 most similar terms to \emph{klonopin} in our vector model. The cosine similarity measure is used to find the closest vectors. The figure also shows that three misspellings for the medication name are among the 20 most similar terms. Interestingly, other medication names appearing closest to \textit{klonopin} (\textit{e.g.}, \textit{ativan} and \textit{xanax}) are typically used for similar purposes (\textit{i.e.}, they are all sedatives). The generic name of the medication (\textit{clonazepam}) also appears close in vector space. The figure verifies that the vector model captures semantic similarities, including misspellings of terms.

\subsection{Gold standard preparation}
We prepared a gold standard consisting of 20 medication names, including trade and generic names. We chose over the counter and prescription medications that have high sales volumes (\emph{e.g.}, \emph{ibuprofen} and \emph{adderall}) so that their correct spellings and misspellings are likely to have high frequencies of occurrences in social media data. We also handpicked medication names that appeared to have interesting properties, which could represent difficult cases for our variant generator. These included medications that have semantic and lexical similarities with other medication names (\emph{e.g.}, \emph{paroxetine} and \emph{fluoxetine}). 

Preparing a gold standard dataset for this task is not trivial. Although, as mentioned, this task has similarities to the task of spelling correction, generating gold standard data for this task is much more complicated than the latter. For spelling correction, the correct spelling for a misspelled term is already known and preparing a dataset simply requires inspecting a set of noisy text and manually mapping the misspellings to their in-vocabulary counterparts \cite{baldwin15}. For the task of variant generation, we have to start with the correct spellings and identify all the possible misspellings from unlabeled data. It is impractical to manually search through millions of social media posts to identify the relevant misspellings and compute their occurrence frequencies. An additional constraint is that the gold standard should only include terms that occur relatively frequently, as including infrequently occurring terms will result in too many variants most of which do not add value to the downstream tasks of data collection and extraction from unlabeled text.

\begin{figure}[htbp] 
	\centering
	\scalebox{0.70}
	{\includegraphics{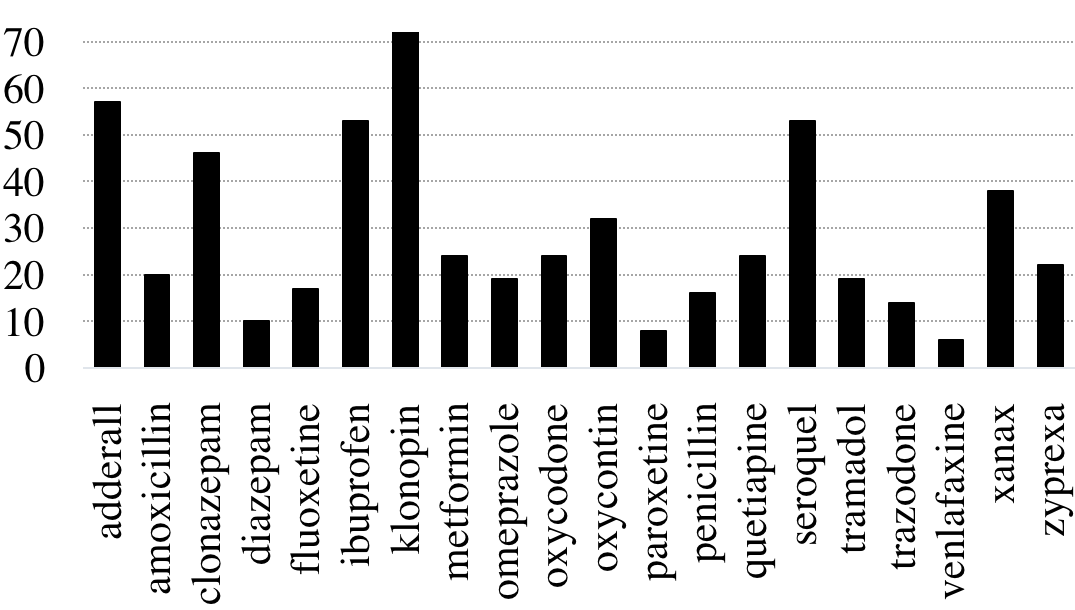}}
	\caption{Distribution of misspellings for the 20 medication keywords.}
	\label{distribs}
\end{figure}

We first performed \textit{fuzzy} searching on an unlabeled dataset consisting of approximately 10 million tweets to identify candidate misspellings for the 20 chosen terms. We included all words  within an edit distance of $min(6,len(keyword)-2)$ from a given medication name. Given two terms \textit{a} and \textit{b}, the edit distance is the minimum number of edit operations that transforms \textit{a} into \textit{b}. These operations include insertion, substitution and deletion of characters. The $min$ function was used because 6 edit distances retrieved too many keywords for short keywords such as \emph{xanax}. We assumed that frequent misspellings would rarely, if at all, be more than 6 edit distances away. To ensure fair comparison, we searched the vocabulary of the language model previously mentioned to generate the candidate misspellings \cite{sarker2017dib} and excluded terms that are absent in the vocabulary based on the assumption that they do not occur frequently enough to be useful. Although it is an extremely time-consuming process, we manually labeled each term in the resulting set to indicate if it is a misspelling or not. This process resulted in 574 true misspellings for the 20 medication names (28.7 per term on average). Figure \ref{distribs} shows the distribution of misspellings per medication.

\begin{figure}[htbp] 
	\centering
	\scalebox{0.30}
	{\includegraphics{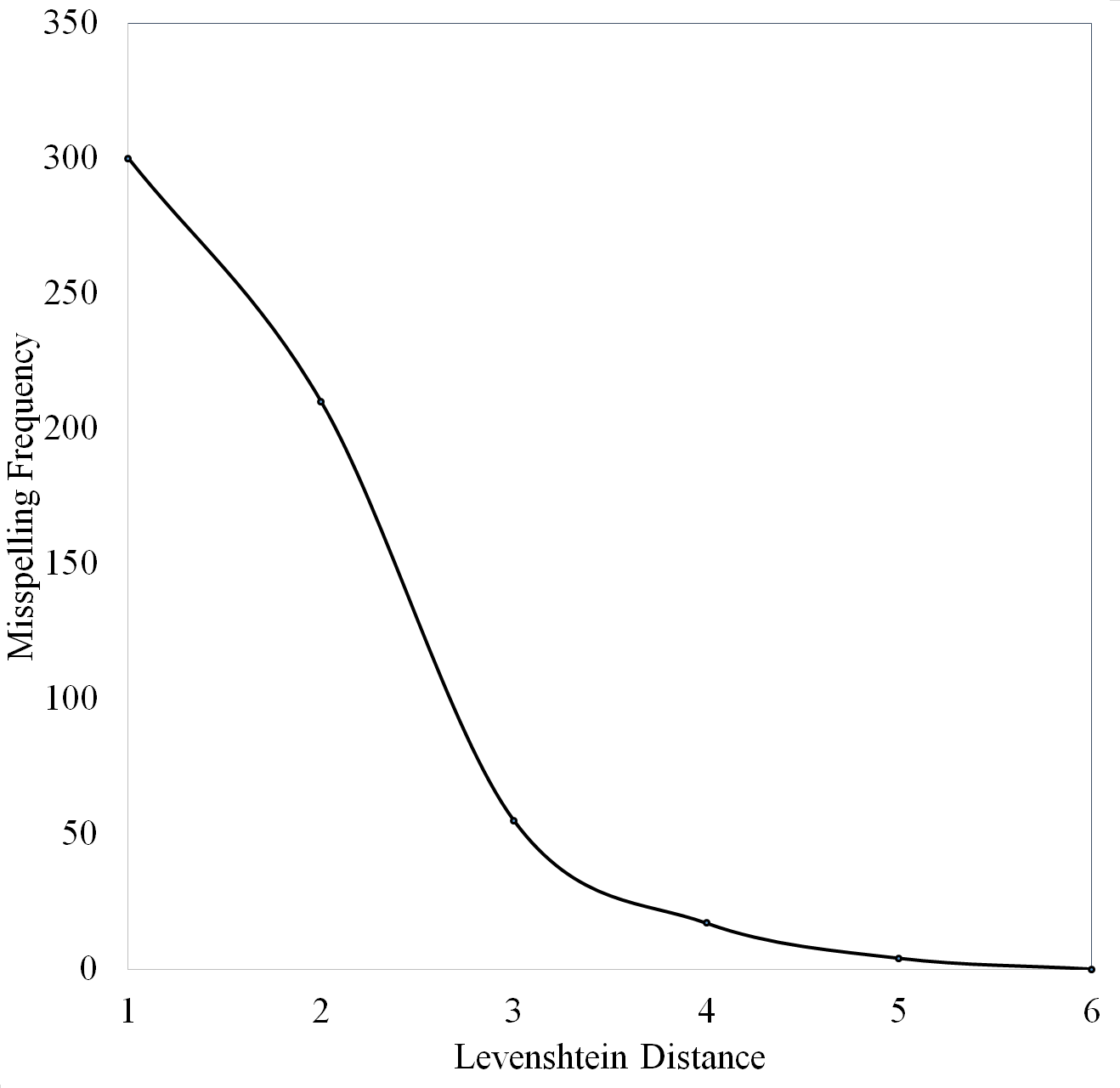}}
	\caption{Distribution of misspelling frequencies against levenshtein distance illustrating that the number of misspellings decrease gradually, with no misspelling occurring beyond a levenshtein distance of 5 (in our dataset).}
	\label{fg3anwe}
\end{figure}

We randomly chose 10 keywords from the set for analysis and algorithm development. We left the remaining 10 for evaluation. We first analyzed the distribution of the frequencies of misspellings against edit distance. Figure \ref{fg3anwe} presents the distribution for the 10 training set medications. The figure illustrates that the frequencies of misspellings and levenshtein distances have an approximately exponential decay-like relationship with no misspellings beyond distance 5, validating our assumption regarding the infrequency of misspellings occurring beyond an edit distance of 6.

\begin{figure}[htbp] 
	\centering
	\scalebox{0.25}
	{\includegraphics{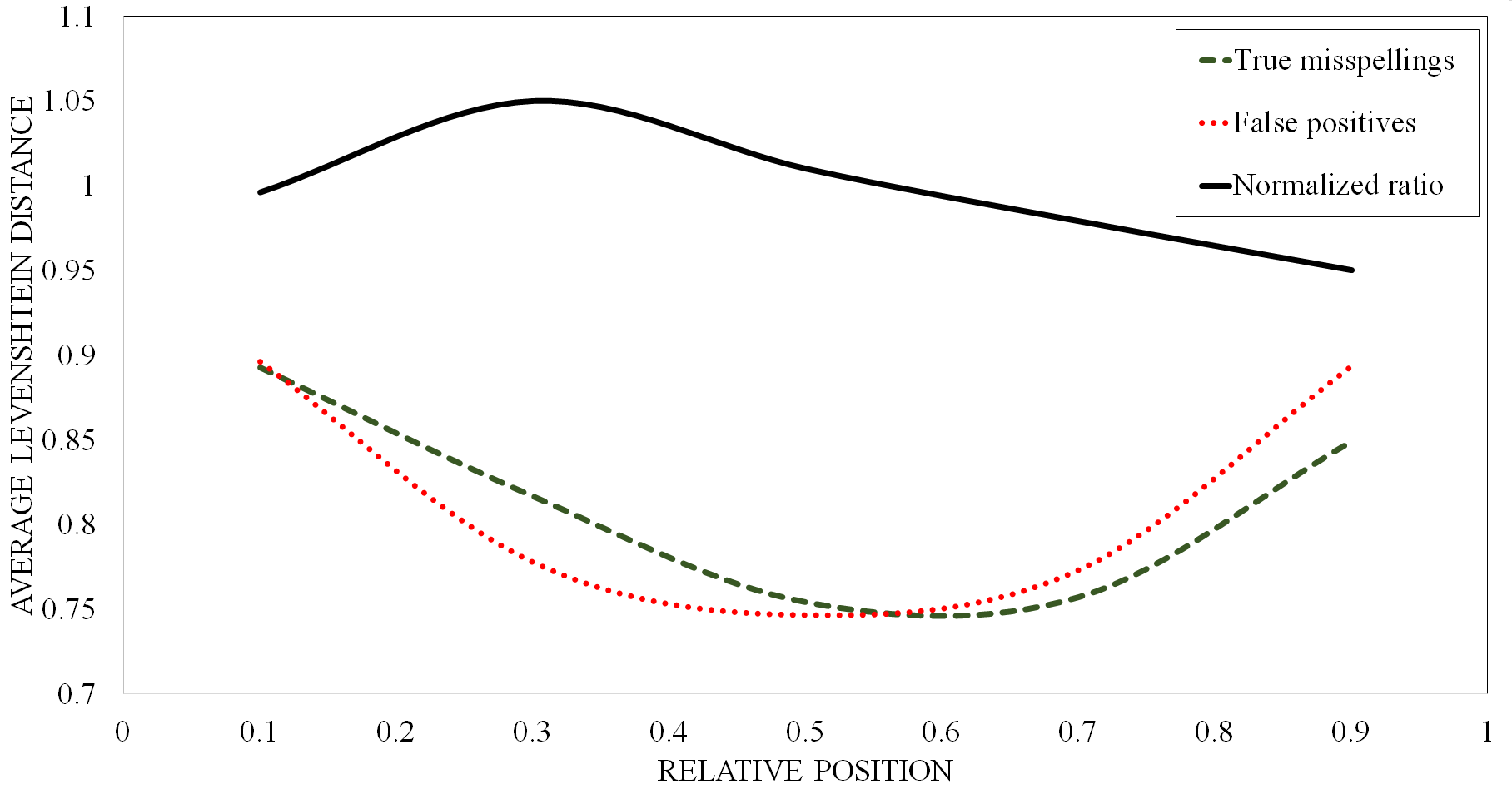}}
	\caption{Distributions of average levenshtein distances between the true misspellings and the false positives. The distribution of the ratios of the two sets are also shown.} 
	\label{fg4new}
\end{figure}

We further inspected the 10 training keywords and their misspellings to analyze how the edit distances were distributed within predefined character windows between the keyword-misspelling pairs. Given a keyword (\emph{e.g.}, \emph{paroxetine}), we wanted to analyze how the lexical similarities were distributed for character windows of length $n$ between the true misspellings of the keyword and non-misspellings that are also within a given edit distance range. To perform this investigation, we passed $n$-character ($2<n\leqslant\frac{len(keyword)}{2}$) windows through each keyword and a corresponding misspelling and computed the edit distance for each window position. We performed the same for non-misspellings. Our intuition was that levenshtein distances may be distributed differently across character sequences between the true misspellings and the false positives that made the initial lexical similarity threshold. We observed that compared to the true misspellings, the non-misspellings tend to have low average levenshtein distance for higher relative character positions. This is because for this specific task, medications belonging to the same class are often identically spelled near the end, reflecting the class of the medication (\emph{e.g.}, \emph{diazepam} and \emph{clonazepam}, \emph{amoxicillin} and \emph{penicillin}). Thus, the analysis suggested that for detecting true misspellings (and hence increasing precision), rewarding lexical similarities among sequences in lower relative position values might be beneficial. Figure \ref{fg4new} shows the distribution of average levenshtein distances at different relative positions for the true misspellings and false positives, along with the normalized ratios of them. We employed a method customized for this property, as discussed later in the paper.  

\subsection{Generating spelling variants}
Algorithm 1 outlines the default system implementation using a stack as the data structure. Given a keyword, the algorithm identifies the $ssl$ semantically closest terms by computing cosine similarities over the vectorized vocabulary. These terms are then passed to a function to compute the levenshtein ratio between the original keyword and the semantically similar terms. Levenshtein ratio is computed using the equation $1 - \frac{ldist}{max(len(a),len(b))}$, where $ldist$ is the levenshtein distance\footnote{The levenshtein distance computation in this case considers substitution to be two edits, while insertion and deletion are considered to be one.} and $max(len(a),len(b))$ is the length of the longer string among the keyword and the candidate misspelling. Each candidate is added to the list of possible misspellings if the levenshtein ratio is above a given threshold ($lt$), and, for each identified misspelling, the same procedure is executed recursively. The levenshtein ratio for a potential misspelling is always computed against the original keyword, ensuring that the algorithm converges once all the terms meeting the levenshtein ratio threshold have been discovered.

\begin{algorithm}[htbp]
	\DontPrintSemicolon
	\SetAlgoLined
	\SetKwProg{Fn}{Function}{}{}
	\SetKwInOut{Input}{Inputs}\SetKwInOut{Output}{Output}
	\Input{\texttt{s} :- the seed keyword/term\\
		\texttt{w} :- the word vector model \\
		\texttt{ssl} :- integer specifying the limit of semantically close terms to search\\
		\texttt{lt} :- the levenshtein ratio threshold; range $[0,1]$ 
	}
	\Output{A set of spelling variants for \texttt{s}}
	\BlankLine
	\Fn{gen\_vars (\texttt{s},\texttt{w},\texttt{ssl},\texttt{lt})}{
		$\texttt{vars}\leftarrow \{\}$ \tcp*{empty dictionary}
		$\texttt{tte}\leftarrow [s]$  \tcp*{stack holding s}  
		$\texttt{aet}\leftarrow [ ]$ \tcp*{empty array}
		\BlankLine
		\While{not \texttt{tte} is \texttt{empty()}}{
			$\texttt{t}\leftarrow \texttt{tte.pop()}$\;
			$\texttt{aet.push(t)}$\;
			$\texttt{sls}\leftarrow \texttt{mostsimilar(w,t,ssl)}$\;
			\ForEach{\texttt{sl} in \texttt{sls}}{
				$\texttt{lss}\leftarrow \texttt{lev\_ratio(sl,s)}$\;
				\If{$\texttt{lss}\eqslantgtr \texttt{lt}$}{
					$\texttt{vars[s]} \stackrel{+}{=} \texttt{lss}$\;
					\If{$! \texttt{lss}$ in $\texttt{aet}$ and $! \texttt{lss}$ in $\texttt{tte}$  }{
						\texttt{tte.push(lss)}	
					}
				}
			}
		}
		\Return \texttt{vars}
	}
	\label{alg11}
	\caption{Spelling variant generation}
	
\end{algorithm}

\subsubsection{Weighted levenshtein ratio}
As discussed, for a given term and a potential misspelling, character sequence similarities in the early parts of the two terms are more likely to be higher for true misspellings. We modified our levenshtein ratio computation by incorporating this information, particularly with the goal of providing a high-precision version of the system. In this system variant, a sliding window of $n$ characters is run through a keyword and a potential misspelling, the levenshtein ratio is computed for the character sequences within the window, and the value is weighted so that the final similarity score is the weighted sum of the individual sequence ratios. Weights are assigned for each bucket of relative positions using the ratio $weight = \frac{fpldist[i]}{tpldist[i]}$, where $i$ is the relative position, and $fpldist$ and $tpldist$ are the distributions of the average levenshtein distances at different relative positions for the true misspellings and non-misspellings, respectively. We used bucket sizes of 0.2, resulting in five weights for all the values within the range $[0,1]$. The weights are normalized by dividing by the median and then scaled, allowing for a reward or penalty of up to $k\%$. This method enables the customization of the variant generation tool to the domain-specific regularities of distinct texts, particularly allowing for a mechanism to reduce the number of false positives. Figure \ref{fg4new} depicts the distributions of average levenshtein distances between the true misspellings and the false positives and the distribution of the ratios of the two sets. In Algorithm 1, the $lev\_ratio()$ function is replaced by its weighted counterpart for this setting. All statistics and parameters are computed and tuned using the development set.

\section{Evaluation and results}
\subsection{Intrinsic evaluation}
We performed intrinsic evaluation of our system using the 10 held-out medication names. We compared the default and weighted approaches of our system to the phonetic variant generation approach proposed by Pimpalkhute et al. \cite{pimpalkhute14}. Based on optimizations over the training set, we used the following parameter settings: $n=5\%$, $lt=0.75$ and $ssl=4000$. The $F_\beta-score$ is used for evaluation and is computed from the precision and recall of the system  as follows:
\begin{equation*}
precision = \frac{tp}{tp + fp} ; recall = \frac{tp}{tp + fn}; F_\beta-score = (1+\beta^2) \times \frac{precision \times recall}{ (\beta^2 \times precision) + recall}
\end{equation*}

where $tp$ is the number of true positives, $fp$ is the number of false positives and $fn$ is the number of false negatives for the system run. The value of $\beta = 1$ is used to compute the $F_1-score$ that gives equal weights to recall and precision, while $\beta = \frac{1}{4}$ is used to compute $F_{\frac{1}{4}}-score$, which provides a higher weight for precision. We included this evaluation metric because for certain health-related data retrieval or concept detection tasks, a more precise version of the system may be preferred over the setting that provides the best $F_1-score$. 

\begin{figure}[htbp] 
	\centering
	\scalebox{0.50}
	{\includegraphics{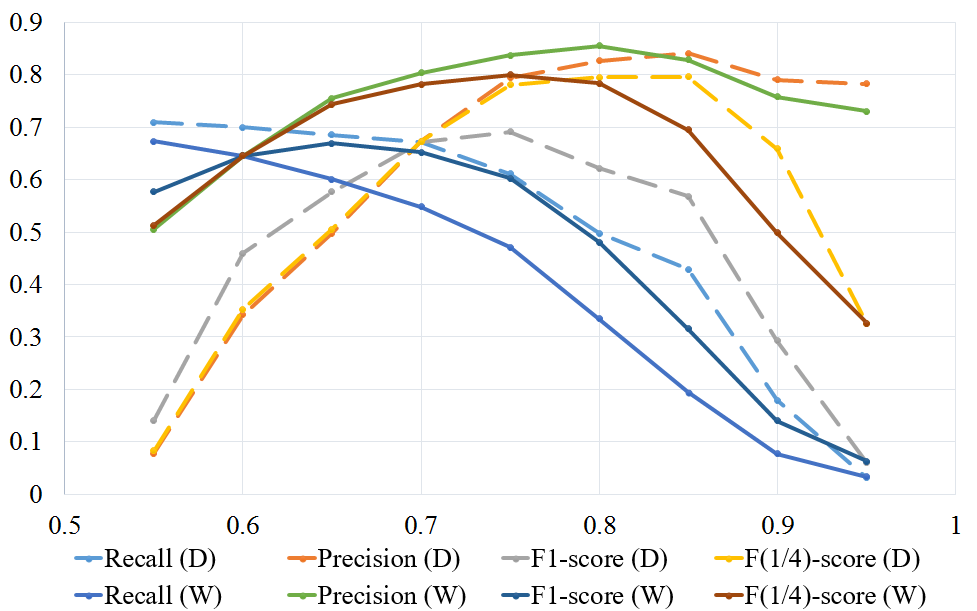}}
	\caption{Recall, precision, F$_1$-score and F$_{\frac{1}{4}}$-score for the two system settings. (D) represents the default version and (W) represents the weighted version.}
	\label{fig3}
\end{figure}

Table \ref{tab1} presents the best results obtained by the systems in terms of F$_\beta$-score on the test set. The default version of our system ($QMisSpell$) obtains the top $F_1-score$ while the weighted version ($QMisSpell_w$) obtains the highest $F_{\frac{1}{4}}-score$. As predicted, the phonetic variant generator achieves very low precision, as it generates many spelling variants that are not relevant. The weighted version of the system achieves the highest precision by marginally outperforming the default system, albeit at some expense to recall. This suggests that the weighted version of the system maybe  suitable for generating smaller numbers of highly precise misspellings. Figure \ref{fig3} illustrates recalls, precisions and F-scores for the two versions of our system perform at different lexical similarity thresholds ($lt$) between 0.55-0.95. It can be observed customization via weighting leads to increases in precision at different thresholds, although recall decreases. The $F_{\frac{1}{4}}-score$, which puts more weight on precision, is also higher for the customized system. The standard approach consistently outperforms the weighted one in terms of overall $F_1$ score, although the differences are small. Table \ref{tab2} presents all the medication keywords and the spelling variants generated by the weighted configuration of our system. False positives generated by the system are underlined.


\begin{table}[htbp]
	\centering
	\begin{tabular}{l c c c c}
		\toprule
		\textbf{System}&\textbf{Recall}&\textbf{Precision}&\textbf{$F_1-Score$}&\textbf{$F_\frac{1}{4}-Score$}\\
		\toprule
		Phonetic&0.49&0.45&0.47&0.45\\
		QMisSpell&\textbf{0.61}&0.79&\textbf{0.69}&0.77\\ 
		QMisSpell$_{w}$&0.47&\textbf{0.84}&0.60&\textbf{0.78}\\

		\bottomrule
		
	\end{tabular}
	
	\caption{Performances for the two variants of our system and the benchmark system. Best score in each column are shown in bold face.}
	\label{tab1}
\end{table}

\begin{longtable}{ll}
		\toprule
		\textbf{Keyword }&\textbf{Generated misspellings}\\
		\toprule
		omeprazole&\shortstack[l]{omaprazole omeprezole omeprizole omeperzole omperazole omeprozole\\ omeprasole omiprazole omeprazol}\\\midrule
		
		paroxetine&\shortstack{paroxotine paroextine paroxitine paroxatine}\\\midrule
		
		klonopin&\shortstack[l]{klonopim klonipin klonepin kolnopin clonopin klomopin klonapins klonoin \\klopopin klonodine klonopan klonopon klonapin klonopn klanopin klenopin \\clonopine klonopins klonopen klonipine klolopin klonopine klonipins klononpin}\\\midrule
		
		diazepam&\shortstack{diazepan diazepams \underline{oxazepam} diazipam diazapam}\\\midrule
		
		fluoxetine&\shortstack{fluoextine fluoxentine fluoxitine fluxoetine  flouxetine fluoxotine fluoxatine duloxetine}\\\midrule
		
		clonazepam&\shortstack[l]{klonazepam clonazpam clonazapam clonazapan clonazipam  clonazepan\\ klonazapam clonezepam clonanzepam clonozepam  clonazepham clonazepem\\ clonazopam clorazepam clanazepam  clonazipan}\\\midrule
		
		zyprexa&\shortstack[l]{zyprea xyprexa xeprexa zyprexia zyprex zypreza zyprexea zyorexa zeprexa zyprxa }\\\midrule
		
		amoxicillin&\shortstack[l]{amoxicillan amoxicilin amoxicillian amoxocillin amoxcillin amoxacillin amoxicllin \\amoxicillion amoxycillin amoxcillian}\\\midrule
		
		tramadol&\shortstack[l]{tramadal tramado tramedol tramadol tramidol tramadoll tramodol tremadol tramamdol}\\\midrule

		xanax&\shortstack[l]{xananx zanax  xanas xanaax xanaxs xanac xanaz}\\\midrule
		oxycontin &\shortstack[l]{oxicontin oxycotins oxycontins oxycottin oxycontin oxycotine oxycintin\\ oxycotin oxycontine}\\\midrule
		
		quetiapine&\shortstack[l]{quatiapine quietapine quetiapin quitiapine}\\\midrule
		
		seroquel &\shortstack[l]{seruquel seraquel seroqule seriquel seroquill seroqel seroquil seroquell\\seroqeul seroqul seroque  seroqual seroguel seroquels seroquelxr serequel}\\\midrule
		
		oxycodone&\shortstack[l]{oxycodine oxicodone oxycoton ocycodone oxycodones oxycodin  oxycodons \\oxycodne oxycondone oxycoden oxycodon oxyxodone oxycodene}\\\midrule
		
		ibuprofen&\shortstack[l]{ibufrofen ibuprofine ibeprofin ibuporfen iubprofin ibeprofen  ibrupofen ibuprofun\\ iboprofen ibuprofrin ibuprofin ibuprofren ibuprofens ibuprofins ibuprohen ibupropen}\\\midrule
		
		venlafaxine &\shortstack[l]{venlafaxin venlaflaxine venlafexine venaflaxine}\\\midrule
		
		adderall &\shortstack[l]{\underline{inderall} addarall adderalls adderrall adderral adderalll adderoll adderal adderallxr}\\\midrule
		
		metformin &\shortstack[l]{metforman metfromin metform metformen metforim metforin metfornin medformin \\metfomin metformine metforming metfirmin  metofrmin }\\\midrule
		
		penicillin &\shortstack[l]{penicillium penicillian penecillin penicillan penicilin  penicillen \underline{ampicillin} \\penicillins penacillin}\\\midrule
		
		trazodone 	&\shortstack[l]{trazidone trozodone trazdone traxodone trazadone trazadones trazodon trazedone \\trazondone trazadon}\\

		\bottomrule

	\caption{Medication keywords and the misspellings generated by the weighted version of our system. False positives generated by the system are underlined.}
		\label{tab2}
\end{longtable}

\subsection{Extrinsic evaluation}
We performed extrinsic evaluation of the system by quantifying the change in retrieval rate (\textit{i.e.}, \% difference in the number of posts retrieved) when variants of a set of original health-related keywords generated by our system were included. To ascertain the generalizability of the system, we used health-related keywords that were not medication names. Specifically, we used five cancer related terms: \textit{carcinoma}, \textit{malignant}, \textit{leukemia}, \textit{metastasis} and \textit{chemotherapy}. We deliberately chose non-medication keywords for this evaluation since these keywords exhibit properties that medication names do not, such as the existence of multiple morphological variations (and their possible misspellings). For example, the term \textit{metastasis} has morphological variants such as \textit{metastasize} and \textit{metastatic}, and their common misspellings, which would be typically included for data collection. Our system generated a total of thirty-three variants, including the original keywords and frequent morphological variants, for the five terms (6.6 variants per keyword on average). 

We queried our in-house Twitter adverse drug reaction database first using the original keywords only and then including the variants.\footnote{Note that this dataset was collected using medication names as keywords, so we only expected a small number of the posts to mention cancer-related terms.} The queried dataset consisted of 7.98 million tweets in total with initially unknown numbers of occurrences of each of these terms. Querying using the original keywords retrieved 5579 tweets. Querying with misspellings along with automatically generated morphological variants retrieved a total of 9348 tweets, an increase of over 67\% in the number of retrieved posts. When morphological variants were excluded and only spelling variants were used, 7677 tweets were retrieved, which represents an increase of 37\% in retrieval rate.

\section{Discussion}
\subsection{Performance}
We have developed a purely data-centric system for generating frequently occurring misspellings and variants for terms that are particularly prone to being incorrectly spelled. The developed system has several advantages over the past phonetic spelling variant generator. Intrinsic evaluations on the 20 medication names showed that the data-centric system significantly outperforms the phonetics-based one \cite{pimpalkhute14}. The proposed system is more precise, as it constrains the variants generated to only those that are semantically close to the original keyword. This ensures that noisy and unrelated variants are not generated by the system. From a practical standpoint, the generation of a small number of common variants, rather than a large number of them, may be crucial since data collection APIs often have restrictions on the total number of keywords that may be used. At the same time, while the phonetic variant generator is restricted to only generating keywords within a levenshtein distance of 1, the proposed system is capable of generating misspellings that are more lexically distant. The brief extrinsic evaluation we performed verifies the usefulness of the system for data retrieval from social media.

\subsection{Error analysis}
We were particularly interested in identifying reasons behind the relatively low recall of our system. Error analysis revealed that approximately 15-20\% of the terms in the gold standard are never retrieved by the $mostsimilar()$ function shown in the algorithm. This suggests that these misspellings are not close in the vector space of the model we used, although they were present in the vector model. The likely reason for this phenomenon is that these keywords did not occur frequently enough for their contexts to be sufficiently interpreted during the embedding generation process. Incorporating a word embedding model built from a larger data set is likely to improve the performance of the system. However, evaluating the effects of different vector models was outside the scope of this study, and we leave that as future work. The weighted version of our system is capable of removing some lexically similar false positives, compared to the default system, leading to the increase in precision. For example, variants \emph{duloxetine} and \emph{paroxetine} are often lexically very similar while also having high semantic similarity (as they belong to the same medication class and are often prescribed for the same conditions, such as depression). The weighted approach, by putting a higher weight at lower relative positions, is capable of removing false positives in many such cases. However, it still generates false positives when two medication names are particularly close both semantically and lexically, as is the case for \textit{oxazepam} and \textit{diazepam},\footnote{Both of these medications belong to the benzodiazepine family.} which are at a levenshtein distance of 2 edits.

\subsection{Usability and customizability}
An important goal for us was to develop a system that is easy to use and customize. Unlike past approaches, our spelling variant generation system does not require any manual steps or external resources (\textit{e.g.}, the Google search API). The algorithm only requires a word embedding model, preferably one that is generated from data that is likely to contain many misspellings for the seed keywords. From an operational perspective, for most practical tasks, the standard setting of the system should suffice and the system can be used in a plug-and-play manner. Additionally, the algorithm does not require any supervised machine learning methods, and, therefore, may be used by medical domain researchers without any prior training in NLP or machine learning. It is, however, possible to incorporate supervision for task-specific customization. We have tested our system, via intrinsic and extrinsic evaluations, on social media data, but we believe that the system will also be useful for data collection and text mining from other noisy data sources such as clinical notes and electronic health records.

The system typically generates a small number of false positives, if any. For practical use, the false positives can be manually removed. The weighted version of the system can be used if greater precision is needed. The weighted version may also be more effective for spelling variant generation problems associated with other topics or when a larger training set is prepared. The weights may also be modified without requiring in-depth programming expertise in Python, which is the programming language used to implement the system. While we used a data-centric approach to customize the weights, they may also be customized via trial and error if needed.

\section{Conclusion}
The first step in conducting health-related NLP research on noisy text sources such as social media and electronic health records typically involves data collection using keyword-based searches. However, noisy text sources invariably contain misspellings, and health domain specific keywords are generally harder to spell. This results in loss of relevant data unless misspellings are generated prior to data collection. While some domain-specific NLP studies have proposed automatic misspelling generation methods, they suffer from various weaknesses. 

In this paper, we proposed a method for the data-centric generation of misspellings. Our proposed method generates semantically similar frequent misspellings for keywords, outperforming the current benchmark spelling variant generation system. The method is simple and fast, and may be used for other restricted-domain data collection tasks from social media and electronic health records. The system may also be customized to reward precision/recall, based on the problem-specific needs. Due to the growing usage of noisy text-based data for research in complex domains and the scarcity of misspelling generation systems, we expect our method to be valuable to the research community. The source code for the system has been made publicly available for the benefit of the health-related text mining and NLP community.

\section*{Acknowledgments}
This work was partially supported by National Institutes of Health (NIH) National Library of Medicine (NLM) grant number NIH NLM 5R01LM011176. The content is solely the responsibility of the authors and does not necessarily represent the official views of the NLM or NIH.



%
%
\bibliographystyle{fullname}
\bibliography{acl2017}

\end{document}